\documentclass[10pt,letterpaper]{article}
\usepackage[top=0.85in,left=2.75in,footskip=0.75in]{geometry}

\usepackage{amsmath,amssymb}

\usepackage{changepage}

\usepackage[utf8x]{inputenc}
\usepackage[section]{placeins}
\usepackage{textcomp,marvosym}
\usepackage{multirow}
\usepackage{cite}

\usepackage{nameref,hyperref}

\usepackage{microtype}
\DisableLigatures[f]{encoding = *, family = * }

\usepackage[table]{xcolor}

\usepackage{array}

\newcolumntype{+}{!{\vrule width 2pt}}

\newlength\savedwidth

\raggedright
\usepackage[aboveskip=1pt,labelfont=bf,labelsep=period,justification=raggedright,singlelinecheck=off]{caption}

\makeatletter
\renewcommand{\@biblabel}[1]{\quad#1.}
\makeatother

\usepackage{lastpage,fancyhdr,graphicx}
\usepackage{epstopdf}
\fancyheadoffset[L]{2.25in}
\fancyfootoffset[L]{2.25in}
\begin{document}
\vspace*{0.2in}

\begin{flushleft}
{\Large
\textbf\newline{Brain Effective Connectome based on fMRI and DTI Data: Bayesian Causal Learning and Assessment} 
}
\newline
\\
Abdolmahdi Bagheri\textsuperscript{1*},
Mahdi Dehshiri\textsuperscript{1},
Yamin Bagheri\textsuperscript{2},
Alireza Akhondi-Asl\textsuperscript{3},
Babak Nadjar Araabi\textsuperscript{1},
\\
\bigskip
\textbf{1} School of Electrical and Computer Engineering, University of Tehran, College of
Engineering, Tehran, Iran
\\
\textbf{2} Department of Psychology, Faculty of Psychology and Education, University of Tehran,
Tehran, Iran
\\
\textbf{3} Department of Anaesthesia, Harvard Medical School, Boston, Massachusetts, USA
\\
\bigskip
%
%





* Abdolmahdibagheri@ut.ac.ir
\end{flushleft}
\section*{Abstract}
Neuroscientific studies aim to find an \textbf{accurate} and \textbf{reliable} brain Effective Connectome (EC). 
Although current EC discovery methods have contributed to our understanding of brain organization, their performances are severely constrained by the short sample size and poor temporal resolution of fMRI data, and high dimensionality of the brain connectome. By leveraging the DTI data as prior knowledge, we introduce two Bayesian causal discovery frameworks -the Bayesian GOLEM (BGOLEM) and Bayesian FGES (BFGES) methods- that offer significantly more accurate and reliable ECs and address the shortcomings of the existing causal discovery methods in discovering ECs based on only fMRI data. Moreover, to numerically assess the improvement in the accuracy of ECs with our method on empirical data, we introduce the Pseudo False Discovery Rate (PFDR) as a new computational accuracy metric for causal discovery in the brain. Through a series of simulation studies on synthetic and hybrid data (combining DTI from the Human Connectome Project (HCP) subjects and synthetic fMRI), we demonstrate the effectiveness of our proposed methods and the reliability of the introduced metric in discovering ECs.
By employing the PFDR metric, we show that our Bayesian methods lead to significantly more \textbf{accurate} results compared to the traditional methods when applied to the Human Connectome Project (HCP) data.
Additionally, we measure the reproducibility of discovered ECs using the Rogers-Tanimoto index for test-retest data and show that our Bayesian methods provide significantly more \textbf{reliable} ECs than traditional methods.
Overall, our study's numerical and visual results highlight the potential for these frameworks to significantly advance our understanding of brain functionality.


\section*{Introduction}
{\textit{Causal connectivity}, also known as the \textit{Effective Connectome} (EC), refers to the causal interaction between different regions  within the brain \cite{friston1994functional}. The EC focuses on understanding how neural activity in one brain region influences the activity in another region, thus revealing the flow of information and causal interactions between different parts of the brain.  Causal discovery of brains mostly focuses on deriving the network based on the fMRI data leveraging the high spatial resolution and abundant data provided by fMRI data.
Targeting a low dimensional network, one can either use techniques like stimulating the specific regions or build models based on physiological understandings, such as Dynamic Causal Modeling  (DCM) \cite{friston2003dynamic}. However, these methods are limited in their ability to perturb only a small number of dimensions among the multitude of regions involved, rendering this type of analysis impractical for high-dimensional ones. As a result, studies rely on \textit{functional connectivity}, which is derived based on the correlation of neuronal activity of the brain regions, such as \cite{Biswal1995Functional, Greicius2003Functional, Power2011Functional, Smith2009Correspondence}. However, the assumption that correlation implies causation is an erroneous approach and the main challenge lies in the disparity between the demand for high-dimensional causal statements and the limitations of providing low-dimensional causal statements.} 

The casual discovery of brain networks faces two main challenges: one with the fundamental impotence of methods and the other with the quality of data. While various computational methods, such as Bayesian network modeling
\cite{ji2016learning}, Granger Causality (GC) \cite{roebroeck2005mapping}, and Structural Equation Models (SEM) \cite{mclntosh1994structural} have been proposed, they tend to fall short from the perspective of accuracy and reliability.
In particular, the Graphical approaches, or the Directed Acyclic Graph (DAG) frameworks, are shown to be more accurate in identifying the underlying causal mechanism based on observational and experimental data \cite{ spirtes2000causation}. 
As the graphical approaches, the Fast Greedy Equivalent Search (FGES) \cite{ramsey2017million}, NOTEARS \cite{zheng2018dags}, and Gradient-based Optimization of DAG-penalized Likelihood for learning linEar DAG Models (GOLEM) \cite{ng2020role} have higher accuracy in large-scale networks, which makes them more appropriate in discovering EC. 
Assuming a static structure of EC, previous studies such as \cite{dubois2020causal} and \cite{zhang2022detecting} demonstrate the effectiveness of the FGES and NOTEARS methods in discovering ECs. However, extracting causality in high-dimensional graphs and a limited number of samples still remains {the main computational challenge}. As shown in \cite{ng2020role}, as the network size increases and the sample size becomes limited, the accuracy of the derived graph significantly decreases.
Aside from computational challenges, methods’ performance can be affected by practical challenges like the low temporal resolution of fMRI, leading to results that are contaminated with fundamental ambiguities such as low reproducibility in discovered connectomes and less accurate results \cite{maier2017challenge,sporns2013human}. 
In \cite{eggeling2019structure}, it is shown that prior information that favors sparsity (such as DTI signals \cite{Tsai2018Reproducibility}) can significantly enhance both the accuracy and reproducibility of discovered graphs when a limited amount of data (such as fMRI signals) is available and the dimension of networks grows (such as brain network).
Therefore, developing Bayesian causal methods based on multimodal data may help overcome the abovementioned challenges.

The DTI technique tracks nerve fibers as they connect potentially functionally associated brain regions. The recent research on multimodal data indicates that these nerve fibers (structural connectivity) constrain the strength and persistence of effective/functional connectivities \cite{lord2012changes,ton2014structure,lee2015adaptation,kang2017bayesian,grimaldi2016anatomical,wang2020multimodal,zamani2022local}. However, the results in \cite{zalesky2012use, goni2014resting} suggest that while there is some degree of overlap between functional and structural connectivity, there are also cases in which they do not align perfectly,i.e., there can be edges in functional connectivity matrix that are not in the binarized structural connectome {($\textrm{SC}_\textrm{B}$)}. The main reason for this misalignment is that functional connectivity refers to the statistical dependencies between different brain regions,  suggesting that the structural connectome does not serve as reliable prior knowledge when exploring functional connectivity.
In contrast, in the context of causal connectivity, there is an alignment between causal connectivity and structural connectivity, that is, the edges in the EC represent a subset of the edges that are present in the $\textrm{SC}_\textrm{B}$ {\cite{sporn2022network}}. It is worth noting that a causal connection between two regions cannot be discovered without the mediation of another variable, assuming that the corresponding edge in $\textrm{SC}_\textrm{B}$ does not exist \cite{Peters2017Elements}.
Additionally, the subcortical regions can potentially act as mediating or confounding variables of cortical regions. 
As a result, including the subcortical regions in the causal discovery process is essential to ensure the satisfaction of the causal sufficiency assumption which is presented in \cite{spirtes2000causation}.
Studies such as \cite{chiang2017bayesian} have shown that developing Bayesian frameworks for discovering ECs and leveraging multimodal data, such as DTI, as prior knowledge improves inference on effective connectivity, compared with unimodal studies.
This research path has garnered significant attention in neuroscience and shows promise in understanding brain organization. Notably, as a causal modeling approach, the DCM method benefits significantly from extracting effective connections with priors on structural connectivity. In \cite{rolls2018effective} and \cite{rolls2020effective}, the ECs for patients with depression and autism, respectively, are extracted by focusing on the structurally connected regions. Specifically, they apply a linearized version of the DCM to only the structurally connected regions, $\textrm{SC}_\textrm{B}$.
In \cite{stephan2009tractography,sokolov2019linking}, the authors presented a Bayesian DCM method using probabilistic structural connectivity, $\textrm{SC}_\textrm{P}$, as prior information. The derived results in these studies are compared with fMRI-based models to illustrate the improvement of the accuracy of the EC model when $\textrm{SC}_\textrm{P}$ is employed as prior information.
Moreover, causality in the brain is investigated in \cite{hinne2014structurally} with the Bayesian partial correlation method, in which the DTI data is approximated with the G-Wishart distribution as the conjugate prior. Likewise, in \cite{crimi2021structurally}, the authors investigate causality by minimizing the reconstruction error of an autoregressive model constrained by the structural connectivity prior.

In this paper, we introduce two new Bayesian causal frameworks, i.e., Bayesian GOLEM (BGOLEM) and Bayesian FGES (BFGES), using DTI data as prior knowledge of the EC discovery to address the mentioned challenges of discovering EC when relying solely on fMRI data. Then, we show that these frameworks provide more accurate and reliable (reproducible) ECs that highlight the potential for these frameworks to advance our understanding of brain function and organization significantly. To compare the accuracy of different methods, we present the Pseudo False Discovery Rate (PFDR) metric based on the False Discovery Rate (FDR) as a computational metric in evaluating the accuracy of discovered ECs.
Then, we measure the reliability of discovered ECs using the Rogers-Tanimoto index for test-retest data.
The main contributions of our work are thus three-folded:
\begin{itemize}
    \item First, we introduce the BGOLEM and BFGES methods as Bayesian causal frameworks to improve the \textbf{accuracy} and \textbf{reliability} of discovering brain effective connectomes while addressing mentioned practical and computational challenges. These methods are Bayesian extensions of the well-established GOLEM and FGES methods.
    \item { Second, we introduce the concept of the Pseudo FDR (PFDR) metric to illustrate the effectiveness of Bayesian causal frameworks, adjust each method's hyperparameters, and compare the accuracy of different methods. Our findings pinpoint the trustworthiness of PFDR as a measurable counterpart of FDR on real data.}
    \item Third, we demonstrate that our Bayesian methods for discovering ECs achieve higher \textbf{accuracy}, as computed using the \textbf{PFDR} metric, and greater \textbf{reliability}, as assessed by the reproducibility metric known as the \textbf{Roger-Tanimoto index}. {The notable PFDR values computed for ECs discovered with non-Bayesian methods with unimodal data depict the weakness of these methods in discovering EC and emphasize the importance of employing Bayesian frameworks with multimodal data that lead to significantly accurate ECs.}
\end{itemize}

\begin{figure}[h!]
\centering
\hspace*{.0 in}
\includegraphics[width=1\textwidth]
{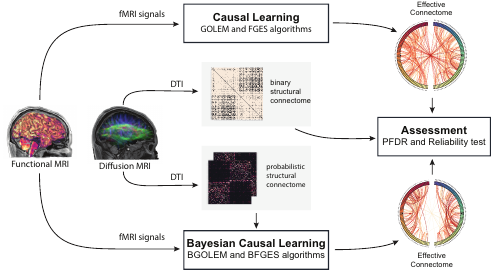}
\caption{Steps in \textbf{Bayesian causal learning} and \textbf{Assessment} of effective connectomes. 
Deriving effective connectomes for both hybrid and empirical data with introduced Bayesian causal frameworks, computing the PFDR value and performing reliability tests. }
\label{fig:mesh1}
\end{figure}

The rest of the paper is organized as follows. The \nameref{2} section presents our Bayesian frameworks, BGOLEM and BFGES methods, and our proposed computational accuracy measure, the PFDR. In the \nameref{3} section, we utilize the developed methods with the hybrid data and illustrate the effectiveness of Bayesian methods by comparing different types of errors. 
Then, we employ Bayesian causal frameworks with HCP data and assess the accuracy and reliability of the discovered ECs through various means. Finally, we discuss the results and limitations of our methods. Fig \ref{fig:mesh1} shows the schematic of our contributions and methods.

\section*{Materials and methods} \label{2}
{In this section, we begin by describing how to compute the probabilistic structural connectome.
Then, we introduce our Bayesian causal frameworks, BGOLEM and BFGES, which are developed based on the GOLEM and FGES methods, respectively. 
In the BGOLEM method, we incorporate prior knowledge into the optimization process of the GOLEM method by masking it on the optimization score. The GOLEM method, which is based on the Gradient-based Optimization of DAG-penalized Likelihood for learning linear DAG Models, provides a foundation for our Bayesian framework.
Similarly, in the BFGES method, we integrate prior knowledge with the Fast Greedy Equivalent Search (FGES) algorithm. In our Bayesian framework, we introduce a modified Bayesian information criterion (BIC) that takes into account the prior knowledge. Then, due to the absence of a gold standard for assessing the accuracy of effective connectivity (EC) methods, we introduce the concept of the PFDR metric as a means to evaluate the performance of the BGOLEM, BFGES, GOLEM, and FGES methods. The PFDR metric is conceptualized with an eye on the idea from the DTI technique and the FDR metric.}
{In the data subsection, we begin by discussing the Causal sufficiency assumption and its relation with our choice for parcellation, explaining its significance and implications for our study. We then outline the reasons for selecting a specific parcellation approach, highlighting its relevance and benefits for our research. Following that, we present the HCP data used in our study, providing details about the data acquisition and preprocessing procedures. Next, we describe the process of generating synthetic and hybrid data, elaborating on the methodologies and considerations involved. We explain how these data sets complement the empirical data that we have gathered for our study.  All the codes for this section including DTI tractography, hybrid data generation, BFGES, and BGOLEM methods are publicly available at} \url{https://github.com/abmbagheri/BGOLEM-and-BFGES}.
\subsection*{Probabilistic structural connectome}
The prior knowledge in the developed Bayesian methods is computed from DTI data. As a result, in order to compute the probabilistic SC, $\textrm{SC}_\textrm{P}$, we use the following formula
\begin{align}
\label{eq:8.1}
\rho^*_{ij}=\frac{\rho_{ij}}{\sum_{j=1}^{j=n}{\rho_{ij}}}
\end{align}
where $\rho_{ij}$ represents the streamlines from seed $i$ to the target region $j$ that is normalized according to the area of the $j^{th}$ region \cite{stephan2009tractography} and $n$ is the number of regions in the brain atlas.

\subsection*{Bayesian causal discovery frameworks}
Consider a parameterized Bayesian network model with $n$ nodes ($B$), defined by $(G, \theta)$, 
where, $G = (V, E)$ is a DAG, $V$ is a set of nodes ($ U=\{X_1, . . . , X_n\}$), $E$ is a set of directed edges (causal relations), and 
$\theta$ is a set of parameters that specify all conditional probability distributions. 
A parameterized Bayesian network $B$ represents and factors a joint distribution over $U$ according to the structure $G$
\begin{align}
\label{eq:1}
P_B(X_1, . . . , X_n)= \prod_{i=1}^n{p(X_i = x_i|Pa^G_i = pa^G_i ,\theta_i)}
\end{align} 
where $Pa^{G}_i$ is the set of parents of $X_i$ in $G$, and $\theta_i \subset \theta$ is the subset of parameters that define the conditional probability of node $X_i\in V$ given its parents in $G$.

\subsubsection*{Proposed Bayesian GOLEM method}\label{211}
The GOLEM algorithm finds a linear DAG model that equivalently represents a set of linear structural equations. 
In GOLEM, $P_B$ in equation \ref{eq:1} is generated by the linear DAG, $X = WX + N$, where, $X=[X_1, . . . , X_n]^T$, $W$ is a $n\times n$ weighted adjacency matrix, and $N = [N_1 ,... , N_n ]^T$ is an independent noise vector.
In \cite{zheng2018dags}, $W$ matrix is found by optimizing a score function $F_B(W,X)$, subject to the structure $G$, for a given samples $X$
\begin{align}
\min_{W\in R^{n\times n} } \quad & F_B(W,X) \nonumber \\
\textrm{s.t.} \quad & G(W)\in DAGs \nonumber \\
F_B(W,X) & =l(W,X)+R_{sparse}(W)
\label{eq:7-2}
\end{align}
where $G(W)$ is the $n$-node graph induced by the weighted adjacency matrix $W$ and $F_B(.) : R^{n\times n}\rightarrow R$ is the score function.
$l(W,X)$ is the maximum likelihood, $R_{sparse}(W)$ is a penalty term that favors sparsity, i.e., having fewer edges. 
The hard optimization problem in equation \ref{eq:7-2} is relaxed into a soft optimization problem presented in \cite{ng2020role} as follows
\begin{align}
\min_{W\in R^{n\times n} } \quad & S_B(W,X) \nonumber \\
S_B(W,X)
&= F_B(W,X)+ R_{DAG}(W) \nonumber \\
&=l(W,X)+R_{sparse}(W)+R_{DAG}(W)
\label{eq:7-41}
\end{align}
where $R_{DAG}(W)$ is a penalty term that favors DAGness of $W$. 
The penalty term for encouraging sparsity is defined as {$\lambda_{\textrm{GOLEM}} \lVert{W} \rVert$.}
The details of the algorithm and proofs are found in \cite{ng2020role}.

\paragraph{\textbf{BGOLEM}}
According to \cite{ng2020role}, the GOLEM method results in lower Structural Hamming Distance (SHD) and FDR values for the sparse structures.
However, for a limited sample size, the accuracy of GOLEM significantly decreases with a higher number of edges and nodes.
To cope with this deficiency, we define the BGOLEM algorithm by suggesting the following sparsity penalty term in \ref{eq:7-41}, to include the prior probability
\begin{align}
\label{eq:7-4}
R_{sparse}(W)={\lambda_{\textrm{BGOLEM}}} \Big\| Q \bigl( P ( G ) ,W \bigl)  \Big\|
\end{align}
where $Q(.): R^{n\times n}\times R^{n\times n} \rightarrow R^{n\times n}$ is a smooth element-wise monotonic function.
This function acts element-wise on the input matrices, such that if a specific element in the $P(G)$ matrix moves in one direction, the corresponding element of the $W$ moves in the same direction. 
Such a sparsity penalty term inhibits/excites an element of the $W$ matrix that has a lower/higher prior probability.
In this paper, we have considered 
\begin{align}
\label{eq:7-5}
Q \bigl( P ( G ) ,W \bigl)=\log\bigl({P(G)}\bigl) \odot {W}
\end{align}
where $\odot$ is the element-wise matrix product.
The choice of the $Q(.)$ function can change the incorporation of prior information in \ref{eq:7-4}.  {Employing the $\log(.)$ function equalizes the skewed distributions, in a highly right-skewed distribution such as the $\textrm{SC}_\textrm{P}$, where most of the values are close to zero, the logarithmic transformation can be beneficial in equalizing the distribution. The impact of extreme values is reduced, which can help bring the distribution closer to a normal or symmetric shape. In the context of incorporating prior knowledge into the final graph, the logarithmic transformation results in a more conservative approach. This conservative incorporation of prior knowledge results in a more balanced representation in the final graph. Moreover, the logarithmic transformation can help reduce the impact of noise on the data. Compressing extreme values mitigates the influence of outliers and makes patterns in the data more apparent.} To find a proper {{$\lambda_{\textrm{BGOLEM}}$}} value, an exhaustive search can be performed to find the optimum $\lambda_{\textrm{BGOLEM}}$ value with the lowest false discovery rate. However, for fMRI data, since there is no access to the ground truth, we use the PFDR metric that is presented in \nameref{22} section to find the optimum hyperparameters, i.e.,$\lambda_{\textrm{BGOLEM}}$, value. 

\subsubsection*{Proposed Bayesian FGES method}\label{212}
The FGES causal discovery method presented in \cite{ramsey2017million}, has two main parts that are: how the algorithm operates and how to define the scoring criteria.
This algorithm has two steps and starts with an empty graph.
In the first step, the forward phase, scores of all alternative one-edge additions to the graph are computed.
The edge that corresponds with the best score is added to the graph.
This process continues until no more improvement is achieved by adding a single edge. 
In the second step, the backward stage, the graph is pruned backward elimination through step-wise single-edge deletion. This procedure continues until any single-edge deletion results in a decline in the score. 
The FGES uses a decomposable score where there is no need to compute the score of the entire graph in each iteration. Only the scores of the nodes with changing parents change. This is the main difference between the FGES and GES methods \cite{chickering2002learning,ramsey2017million}.
The decomposable scoring criteria $S_B$ \cite{ramsey2017million}, is defined by 
\begin{align}
\label{eq:2}
S_B(G,D) =\sum_{i=1}^n{s(X_i,Pa^G_i)}
\end{align} 
where $D$ is a set of observed data and $s(X_i,Pa^G_i)$ is the score corresponding to a node $X_i$ and its parents $Pa^G_i$.

The Bayesian score function $S_B$ using the relative log posterior of $G$ is
\begin{align}
\label{eq:3}
S_B(G,D)=\log{p(G)}+\log{p(D|G,\theta)}
\end{align} 
where $p(G)$ is the prior probability of $G$, and $p(D|G)$ is the conditional likelihood of $D$ \cite{ramsey2017million}
In particular, in \cite{haughton1988choice}, the authors show that equation \ref{eq:3} for curved exponential models can be approximated using Laplace’s method for integrals, yielding
\begin{align}
\label{eq:4}
S_B(G,D)\approx \log{p(G)}+\log{p(D|G,\hat{\theta})}-\frac{d}{2}\log(m)
\end{align} 
where $\hat{\theta}$ denotes the maximum-likelihood estimate for the network parameters, $d$ denotes the dimension (i.e., number of free parameters) of $G$, and $m$ is the number of records in $D$. In \cite{ramsey2017million}, the decomposable score for the FGES method is defined by the sum of the second and third terms in equation \ref{eq:4} that is also known as the BIC, i.e.,
\begin{align}
\label{eq:4-1}
S_B(G,D)=\sum_{i=1}^n{\bigl(\log{p(X_i|\hat{\theta_i},Pa_i^G)}-\lambda_{\textrm{FGES}}\frac{d_i}{2}\log(m)\bigl)}
\end{align} 
where {$\lambda_{\textrm{FGES}}$} is the hyperparameter of the BIC score.
\paragraph{\textbf{BFGES}}
Similar to the GOLEM method, the FGES method has difficulties discovering DAGs with a higher number of nodes and edges, and a limited number of samples, as illustrated in \cite{zheng2018dags}.
While the $\log{p(G)}$ term is neglected in the BIC scoring \ref{eq:4-1}, this term can play a dominant role in model selection of large-scale graphs \cite{eggeling2019structure}.
In our proposed Bayesian FGES algorithm, considering the equations \ref{eq:2} and \ref{eq:3}, we derive the decomposable version of equation \ref{eq:4} as follows
\begin{equation}
\begin{split}
\label{eq:5}
S_B(G,D)& =\log{p(G)} +\sum_{i=1}^n{s_B(X_i,Pa^G_i)}\\
& =\log{p(G)}+
\sum_{i=1}^n{\bigl(\log{p(X_i|\hat{\theta_i},Pa_i^G)}-\lambda_{\textrm{BFGES}}\frac{d_i}{2}\log(m)\bigl)}
\end{split}
\end{equation}
where $d_i$ denotes the number of parameters for the structure of $X_i$ and its parents. The procedure of deriving equation \ref{eq:5} is quite similar to that of equation \ref{eq:4-1}, but without ignoring the  prior probability.
If we assume a uniform prior for the existence of an edge, all three probabilities of $X_i \leftarrow X_j$, $X_i \rightarrow X_j$ and $X_i \dots X_j$ ($X_i$ and $X_j$ are not connected) are equal to $\frac{1}{3}$.
Since the probability of $X_i$ not being a parent of “$X_j$ is equal to $\frac{2}{3}$", then ${p(G)}= {\frac{1}{3}}^{N_G}\times {\frac{2}{3}}^{\bigl(\frac{(n^2-n)}{2}-N_{G}\bigl)}$, where $N_{G}$ an $n$ are the number of edges and number of nodes in graph $G$, respectively.
Similar to the FGES algorithm, in the BFGES algorithm, we start with an empty graph, which means that $p(G)$ is the probability of a graph $G$ with no edges.
As a result, the algorithm starts with 
\begin{equation}
\begin{split}
\label{eq:5_1}
p(G)= \prod_{i=1}^n{\prod_{j=1}^n{\bigl(1-p(X_i\leftrightarrow X_j})\bigl)}
\end{split}
\end{equation}
where $ p(X_i\leftrightarrow X_j)= p(X_i\rightarrow X_j)+ p(X_i\leftarrow X_j)$ is the edge existence probability between nodes $i$ and $j$, $i\neq j$.
Consequently, effect of adding an edge from node $i$ to node $j$ is to subtract $\log \bigl(1-p(X_i\leftrightarrow X_j)\bigl)$ from the score and add $\log \frac{1}{2}p(X_i\leftrightarrow X_j)$ to the score 
\begin{equation}
\begin{split}
\label{eq:5_2}
S_B(G^{'},D)= S_B(G,D)-\log \bigl(1-p(X_i\leftrightarrow X_j)\bigl)+\log \frac{1}{2}p(X_i\leftrightarrow X_j)
\end{split}
\end{equation}
where in this equation, $S_B(G^{'},D)$ is the score of graph $G^{'}$ that is derived from adding $X_i\rightarrow X_j$ to graph $G$.

In the backward stage of the BFGES algorithm, the score of the derived graph is subtracted with $\log \frac{1}{2}p(X_i\leftrightarrow X_j)$ and $\log \bigl(1-p(X_i\rightarrow X_j)\bigl)$ is added when the edge from node $j$ to $i$ is eliminated
\begin{equation}
\begin{split}
\label{eq:5_3}
S_B(G^{'},D)= S_B(G,D)+\log \bigl(1-p(X_i\leftrightarrow X_j)\bigl)-\log \frac{1}{2}p(X_i\leftrightarrow X_j)
\end{split}
\end{equation}
where $S_B(G^{'},D)$ is the score of graph $G^{'}$ that is derived from removing $X_i\rightarrow X_j$ from graph $G$.

In sum, the BFGES algorithm runs similarly to the FGES algorithm with the modified scoring \ref{eq:5} to incorporate prior knowledge in causal discovery. Similar to the BGOLEM method, for fMRI data, we use the PFDR metric (\nameref{22} section) to find the optimum {$\lambda_{\textrm{BFGES}}$} value. 

\subsection*{The PFDR metric}\label{22}
To measure the reliability of methods in discovering causality, False Discovery Rate (FDR) is a powerful full metric to assess the accuracy of discovered edges, which is defined as follows
\begin{align}
\label{eq:7.01}
\textrm{FDR}=\frac{FP}{NoDE}
\end{align}
where $FP$ and $NoDE$ are the number of False Positives and the Total Number of Discovered Edges, in the true underlying EC, respectively \cite{benjamini1995controlling}. In the assessment of discovered ECs, there is no access to the true causal underlying graph. As a result, computing $FP$ and consequently, FDR value is not possible.

In the DTI technique, the relative diffusivity of water in a voxel into directional components is quantified. 
The longest axis of the diffusion ellipsoid, which is estimated based on diffusion tensors, is used to track nerve fibers as they travel between potentially functionally associated brain regions \cite{huettel2004functional}. 
The absence of an edge in $\textrm{SC}_\textrm{B}$ indicates that there is no physical connection between two regions which implies that no effective connection is possible.
On the other hand, the existence of an edge in this matrix implies a physical path between two regions, which opens the door for effective connectivity.
Therefore, from the absence of an edge in this matrix, one can infer that the corresponding element of the EC must be zero. The presence of an edge in $\textrm{SC}_\textrm{B}$ allows for corresponding non-zero elements in EC.
We exploit DTI-based $\textrm{SC}_\textrm{B}$ information and propose the Pseudo FDR (PFDR) metric, which is defined as follows
\begin{align}
\label{eq:7.1}
\textrm{PFDR}=\frac{FP^{'}}{NoDE}=\frac{FP^{'}}{{FP^{'}}+TP^{'}}
\end{align}
where $FP^{'}$ is the number of edges that are one in EC and zero in $\textrm{SC}_\textrm{B}$ and  $TP^{'}$ is the number of edges that are one in both EC and $\textrm{SC}_\textrm{B}$.
From the definition of $FP$ and $FP^{'}$, false positives in PFDR computations are a subset of true false positives in FDR computation, as a result, $FP^{'} \leq FP$. Note that PFDR and FDR can be equivalent in the case the undirected causal graph of the underlying brain's causal mechanism exactly matches the structural graph.
Contrary to the FDR metric, computing the PFDR metric is practicable, while we do not know the true underlying graph. 
This metric enables us to measure and compare the accuracy of the ECs discovered with causal discovery methods. The PFDR can be used to adjust the hyperparameters of causal discovery methods, as well. 
In the ~\nameref{312} section, we experimentally explore the relationship between FDR and PFDR by deriving both values for hybrid data and computing their correlation. {This analysis aims to emphasize that these two metrics can be used interchangeably.}
Details and discussions on Pseudo FDR can be found in Appendix 1.

\subsection*{Data}
\paragraph{Casual sufficiency and selected Parcellation}
The causal sufficiency assumption is the concept that addresses the question that whether any extra variable need to be considered in the process of the causal discovery process of a causal model \cite{spirtes2000causation}.  This concept refers to the condition where, if a set of variables $X$ are causally sufficient, it should not omit any hidden common cause $C \notin X$ that has the ability to influence more than one variable within $X$ \cite{Spirtes2010Introduction}.
Since the subcortical regions receive information flow originating from the brain stem, they have the potential to act as confounding variables to the cortical regions and may play a role as confounding variables in their causal relationships.
As a result, to ensure that the causal sufficiency assumption is satisfied, the subcortical regions are included in our analysis.
Furthermore, consider the scenario where region A comprises sectors $a_1$ and $a_2$, while region B consists of sectors $b_1$ and $b_2$. In this case, sector $a_1$ might exert a causal influence on sector $b_1$, while sector $b_2$ could have a causal influence on sector $a_2$. Such interdependencies between sectors can pose challenges during the causal discovery process and potentially disrupt the identification of causal relationships.
To mitigate these challenges, it becomes crucial to select an atlas that encompasses a rich array of brain regions. By utilizing an atlas with a greater number of distinct regions, we increase the likelihood of capturing the individual sectors involved in the causal relationships accurately.
Correspondingly, we employ the Destrieux atlas \cite{destrieux2010automatic} that satisfies our concerns. The List of ROIs of the Destrieux atlas is presented in Appendix 2.
\paragraph{{Data acquisition}} In this study, we used DWI and {resting-state fMRI} data from unrelated subjects of the “HCP1200” data set (March 2017 data release of healthy adults aged 22–35) \cite{van2013wu}. 
The HCP data sets were acquired using protocols approved by the Washington University institutional review board, and written informed consent was obtained from all subjects. All the authors have been approved by HCP to use this data.
\paragraph{{DTI data preprocessing}}
DWI images were acquired using a 3T ‘Connectome Skyra’, provided with a Siemens SC72 gradient coil and stronger gradient power supply with maximum gradient amplitude (Gmax) of 100 mT /m (initially 70 mT /m and 84 mT /m in the pilot phase) sequence and 90 gradient directions equally distributed over 3 shells (b-values 1000, 2000, 3000 mm/s 2 ) with 1.25mm isotropic voxels \cite{sotiropoulos2013advances}.
The preprocessing steps included distortion and motion correction, brain extraction, and eddy correction for diffusion images, as well as registration of structural and diffusion data to each other in native space \cite{Glasser2016the}.
In this study, we employed probabilistic tractography that considers multiple possible pathways at each voxel to find any possible structural connection. These preprocessing steps were already performed in the downloaded package from
\href{https://www.humanconnectome.org/study/hcp-young-adult}{HCP web page}.
The MRtrix3 software is used to perform whole-brain probabilistic tractography, which involved using the 5TTgen function with the FSL option to segment subcortical nuclei in T1-weighted images. The segmented subcortical ROIs are then combined with cortical ROIs from the Desikan-Killiany atlas that are segmented by Freesurfer during the preprocessing steps in the HCP pipeline. The $msmt_5tt$ algorithm is used to estimate the response function for diffusion images, and the iFOD2 algorithm is used with the multi-tissue multi-shell -$msmt_csd$ flag to generate FOD (fiber orientation density) images. The tckgen algorithm is then used to generate 50 million streamlines from the FOD images with a maximum tract length of 250 mm to obtain whole-brain probabilistic tractography. Finally, the -tck2connectome function is employed to find the number of streamlines that started from each ROI and ended in all other ROIs to determine the structural connectivity between all nodes.

\subsubsection*{Synthetic and hybrid data}

The synthetic data, the first group of data, is generated using a random graph generated from the scale-free (SF) model, as detailed in \cite{zheng2018dags}. This process involves generating graphs with a degree of 7 and varying the number of nodes to be $50, 75, 100, 125,$ and $150$, along with 300 time points.  For each number of nodes, a total of 30 datasets are generated. This group of data is employed to compare the results and effectiveness of our Bayesian methods with the results in \cite{ng2020role}.

To generate hybrid data, the second group of data, two main steps are involved. The first step is to generate random causal graphs based on the DTI data from the HCP. These causal graphs represent the underlying DAG that defines the causal connectivity between different brain regions.
Once the causal graphs are established, the second step is to generate fMRI data based on these causal graphs and the hemodynamic response function. This involves simulating the neural activity within the brain regions defined by each causal graph and then convolving this neural activity with the hemodynamic response function to generate the corresponding fMRI signals.
We create 50 causal graphs with 164 nodes and the final hybrid data consists of 1200 time points. 
The hybrid data is used to show the validity of the PFDR metric.

More in-depth information on the generation of Hybrid data can be found in Appendix 1.
\subsubsection*{Empirical data}
The HCP MRI data acquisition protocols and procedures have previously been described in full detail \cite{smith2013resting}.
We used the minimally preprocessed images of \textit{resting state fMRI}, `rsfMRI', and \textit{Diffusion Tensor Imaging (DTI)}, `dMRI', for 50 unrelated subjects that were provided by the HCP S1200 Release. fMRI resting-state runs (HCP file name: rfMRI-REST1 and rfMRI-REST2) were acquired in four separate sessions on two different days, with two different acquisitions (left to right, or LR, and right to left, or RL) per day \cite{glasser2013minimal}. 
To eliminate any potential systematic bias or artifacts that may merge due to the acquisition process, one can average the two sessions of each day \cite{Physiological2011Triantafyllou} or average the results of applying any connectome discovery methods to two sessions \cite{Conti2019Variability}. In our study, for each subject, the mean of the first day's two sessions is used as test data (Mean 1), while the second day's two sessions are used as retest data (Mean 2).
Averaging allows for a comprehensive analysis that incorporates data from both acquisition directions, enhancing the reliability and validity of the results.
Note that in both empirical and hybrid data, we use DTI data of the same subjects.

\section*{Results}\label{3}
Applying the developed Bayesian methods, the probabilistic $\textrm{SC}_\textrm{P}$ is computed using equation \ref{eq:8.1}. To obtain the binary structural connectome ($\textrm{SC}_\textrm{B}$) that is used in the PFDR metric, in the first step, the number of tracts for each subject is binarized using a threshold of 50$\%$ of stream counts. In the second step, majority voting is applied to the results of the first step for 50 subjects to compute the final binary SC, $\textrm{SC}_\textrm{B}$. 
It is worth noting that studies such as \cite{Tsai2018Reproducibility} have shown that it is reasonable to exclude at least $80\%$ of the edges when deriving the binary $\textrm{SC}_\textrm{B}$.
However, in our study, we have opted to use a $50\%$ threshold for binarization. This choice is motivated by the significance we attribute to zero values in the $\textrm{SC}_\textrm{B}$ used in the PFDR metric, as well as our intention to be more conservative in excluding edges that are not consistently connected across our subjects. By employing majority voting on zero values, we aim to ensure that the derived $\textrm{SC}_\textrm{B}$ is as conservative as possible in capturing the absence of connections.
It is worth noting that extracted $\textrm{SC}_\textrm{B}$ has $33\%$ zero elements.

In the following, to highlight the effectiveness of our Bayesian methods, we apply the GOLEM and FGES methods along with the BGOLEM and BFGES on the synthetic data (SF7) and show the results in terms of the FDR (False Discovery Rate) and the Percentage of Total Errors. The Percentage of Total Errors is defined as follows,
\begin{align}
\frac{\textrm{ False\ Positives }+\textrm{False\ Negatives}}{ \textrm{Total\ Number\ of\ Discovered\ Edges} }
\end{align}
Then, to assess the reliability and validity of the proposed PFDR metric,
we compute the correlation of the FDR and PFDR values of the ECs discovered by applying FGES, BFGES, GOLEM, and BGOLEM to the hybrid data.
To apply the Bayesian causal frameworks to the empirical data, we utilize the $\textrm{SC}_\textrm{P}$ of each subject as prior knowledge for the EC discovery process of the same subject. Results for these data using all four methods are obtained by adjusting their hyperparameters based on the minimum PFDR values observed in the test data.
In comparing the accuracy of the Bayesian method with non-Bayesian methods, to avoid any biases, the PFDR for each subject is computed using the $\textrm{SC}_\textrm{B}$, which is obtained by excluding the same subject from a group of 50 subjects. In the end, to test the reliability of the ECs discovered using our Bayesian methods is examined using the Roger-Tanimoto index, and the results are compared with those obtained from the ECs discovered using the FGES and GOLEM methods. Visual comparisons of the ECs discovered with different methods are also provided to facilitate a comprehensive analysis.

\subsection*{Results for the synthetic and hybrid data}
In this subsection, we present a comparative analysis of the results and effectiveness of our Bayesian methods in contrast to the outcomes obtained using the traditional method with synthetic data. For Hybrid data, both the FDR and PFDR values are computable. Subsequently, by comparing the PFDR and FDR values of applying both Bayesian and non-Bayesian methods to hybrid data, we evaluate the trustworthiness and validity of the PFDR metric. As a result, the optimal values of $\lambda_{\textrm{FGES}}$, $\lambda_{\textrm{GOLEM}}$, $\lambda_{\textrm{BFGES}}$ and $\lambda_{\textrm{BGOLEM}}$ are determined based on minimizing the Percentage of Total Errors, as ground truths are available for these two groups of data.

\begin{figure}[!htb]
\centering
\hspace*{-.0 in}
\includegraphics[width=1\textwidth]{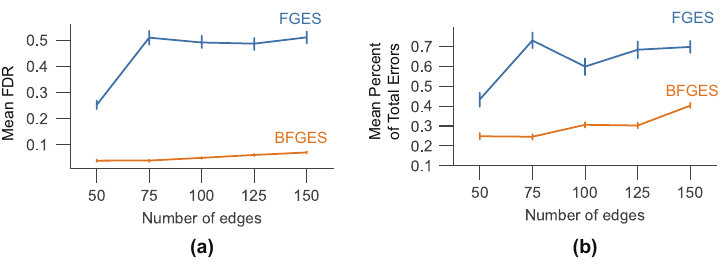}
\caption{ The mean FDR and the mean Percent of Total Errors for graphs with different number edges.
\textbf{(a)}: The mean FDR for the FGES and BFGES methods
\textbf{(b)}: The mean Percent of Total Errors for the FGES and BFGES methods
}
\label{fig:mesh3_0}
\end{figure}

\begin{figure}[!htb]
\centering
\hspace*{-.0 in}
\includegraphics[width=1\textwidth]{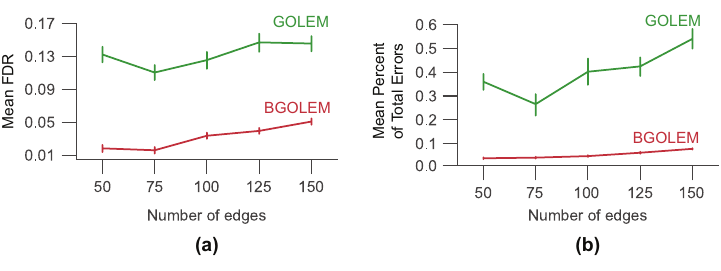}
\caption{ The mean FDR and the mean Percent of Total Errors for graphs with different number edges.
\textbf{(a)}: The mean FDR for the GOLEM and BGOLEM methods
\textbf{(b)}: The mean Percent of Total Errors for the GOLEM and BGOLEM methods
}
\label{fig:mesh3_001}
\end{figure}

The results obtained from applying the FGES and BFGES methods, along with the GOLEM and BGOLEM methods, on synthetic data with varying numbers of nodes are depicted in Fig \ref{fig:mesh3_0} and \ref{fig:mesh3_001}.

Fig \ref{fig:mesh3_0} presents the mean FDR and mean Percentage of Total Errors for both the FGES and BFGES methods. 

Fig \ref{fig:mesh3_001} presents the mean FDR and mean Percentage of Total Errors for both the GOLEM and BGOLEM methods. The hyperparameter of each casual discovery method is determined based on the minimum Percentage of Total Error.

\begin{figure}[!htb]
\centering
\hspace*{-.0 in}
\includegraphics[width=1\textwidth]{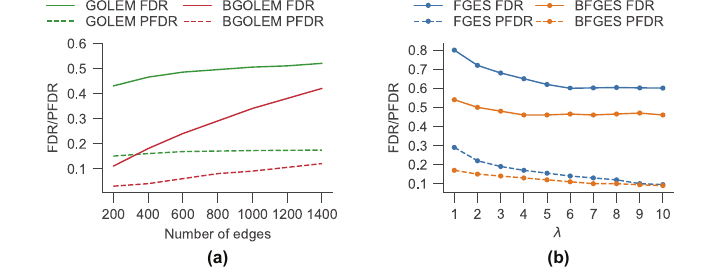}
\caption{ Dependency of the PFDR and FDR metrics for ECs derived with Bayesian and Non-Bayesian methods on hybrid data with 164 nodes. \textbf{(a)}: The correlation coefficients ($R$) of PFDR and FDR values for the ECs of the GOLEM and BGOLEM are $98.9\%$ and $99.9\%$, respectively \textbf{(b)}: The correlation coefficients of PFDR and FDR values for FGES and BFGES are $80.9\%$ and $97.7\%$, respectively.}
\label{fig:mesh3_1}
\end{figure}

In Fig \ref{fig:mesh3_1}, we illustrate the computed FDR and PFDR values for the ECs derived using both Bayesian and non-Bayesian frameworks on hybrid fMRI data consisting of 164 nodes.
In Fig \ref{fig:mesh3_1}-a, the FDR and PFDR values of ECs are presented for the GOLEM and BGOLEM methods, considering different thresholds on the number of edges. Furthermore, the correlation coefficients ($R$) between the PFDR and FDR values for the ECs obtained using the GOLEM and BGOLEM methods are $98.9\%$ and $99.9\%$, respectively.
The FDR and PFDR values for the ECs discovered with the FGES and BFGES methods, considering different values of $\lambda_{\textrm{FGES}}, \lambda_{\textrm{BFGES}}$, are displayed in Fig \ref{fig:mesh3_1}-b. The correlation coefficients between the PFDR and FDR values for the ECs derived from the FGES and BFGES methods are $80.9\%$ and $97.7\%$, respectively. 

\subsection*{Results for the empirical data}\label{312}
In this section, we present and compare the results of applying our methods and traditional methods to empirical data, both numerically and visually.

\begin{figure}[h!]
\centering
\includegraphics[width=\textwidth]{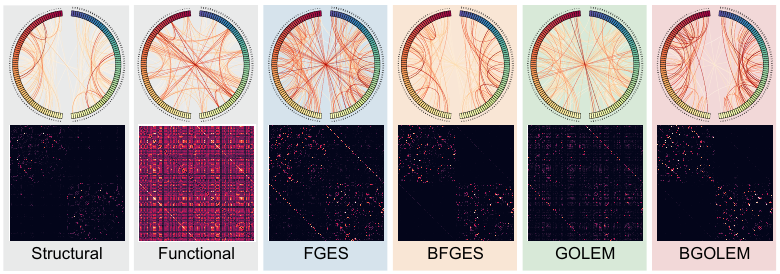}
\caption{Comparison of Structural connectome, functional connectome and ECs discovered with the FGES, BFGES, GOLEM and BGOLEM methods. The first row compares the symmetry of structural connectomes, correlation-based, and ECs and edges between two hemispheres of each connectome. The second row, compares the results of ECs discovered with $\textrm{SC}_\textrm{P}$, functional connectome, and  ECs of non-Bayesian  methods. 
}
\label{fig:mesh7_0}
\end{figure}

Fig \ref{fig:mesh7_0} presents the discovered effective connections (ECs) using the FGES, BFGES, GOLEM, and BGOLEM methods for 148 regions encompassing the left and right hemispheres. For clarity, this figure excludes 16 subcortical regions, focusing on illustrating the edges between the two hemispheres. Additionally, the figure includes the structural connectome and functional connectome, allowing for a comprehensive graphical comparison of these connectomes with the discovered ECs

To numerically investigate the \textbf{accuracy} and \textbf{reliability} of the ECs shown in Fig \ref{fig:mesh7_0}, first, we employ the \textbf{PFDR} as an accuracy metric and then, we use the \textbf{Roger Tanimoto} index as a reproducibility metric.

\begin{figure}[h!]
\centering
\hspace*{-.0 in}
\includegraphics[width=1\textwidth]{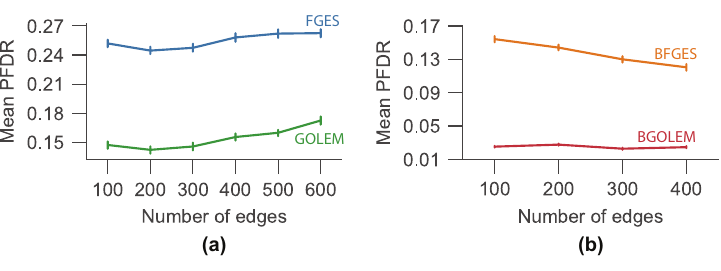}
\caption{The PFDRs for the ECs discovered with the GOLEM, FGES, BGOLEM, and BFGES methods. \textbf{(a)}: The PFDR for ECs with the best hyperparameters of the GOLEM and FGES methods \textbf{(b)}: The PFDR of ECs with the best hyperparameters of BGOLEM and BFGES methods}
\label{fig:mesh6}
\end{figure}

The mean PFDRs for different numbers of edges are presented for the FGES, GOLEM, BFGES and BGOLEM methods in Fig \ref{fig:mesh6}.
In Fig \ref{fig:mesh6}.a, the PFDR values for ECs discovered with the FGES and GOLEM methods are shown, for different numbers of edges.  The lowest mean PFDR value for the FGES method is $24.6\%$ for the penalty coefficient between $4$ and $6$. The lowest value of the mean PFDR of the GOLEM method is between $[12,17.2]\%$ for the different number of edges and variation of $\lambda_{\textrm{GOLEM}}$. The mean PFDR values for ECs of the BGOLEM and BFGES methods are derived with respect to their lowest mean PFDRs and shown in Fig \ref{fig:mesh6}.b. The PFDR values of ECs are $[12.1-15.2]\%$ and $[3-3.3]\%$ for BFGES and BGOLEM methods, respectively.
More analyses and results on empirical data, the schematic of $\textrm{SC}_\textrm{B}$, $\textrm{SC}_\textrm{P}$ and the effect of hyperparameter variation on the PFDR of empirical data can be found in Appendix 1.

In Table \ref{table:1}, we compare the PFDR values for the Mean 1 (test) and Mean 2 (retest) data when 200 edges are selected. 
To assess the significance of employing our Bayesian methods compared to the non-Bayesian method in terms of \textbf{accuracy}, we conducted the Paired T-test on the computed PFDR values. This test is selected to be performed on the PFDR values due to appropriation of this test for paired observations that the differences between pairs are normally distributed.
The p-value obtained for comparing the PFDR values of BFGES and FGES is less than 1e{-3}, indicating a significant difference between these two methods. Similarly, the p-value for comparing the PFDR values of BGOLEM and GOLEM is also less than 1e{-3}, indicating a significant difference between these two methods as well.

\begin{table}[h!]
\centering
\caption{Results of applying Bayesian methods on mean PFDR values for ECs with 200 edges for test and retest data }
\begin{tabular}{ccc}
\label{table:1}
\cellcolor[HTML]{C0C0C0}                                  & \multicolumn{2}{c}{\cellcolor[HTML]{C0C0C0}\textbf{Data}}                         \\ \cline{2-3} 
\multirow{-2}{*}{\cellcolor[HTML]{C0C0C0}\textbf{Method}} & Mean 1                                 & Mean 2                                  \\ \hline
\multicolumn{1}{|c|}{FGES}                                & \multicolumn{1}{c|}{$0.242 \pm 0.0035$} & \multicolumn{1}{c|}{$0.254 \pm 0.0042$} \\ \hline
\multicolumn{1}{|c|}{BFGES}                               & \multicolumn{1}{c|}{$0.142 \pm 0.0015$} & \multicolumn{1}{c|}{$0.154 \pm 0.0022$} \\ \hline
\multicolumn{1}{|c|}{GOLEM}                               & \multicolumn{1}{c|}{$0.141 \pm 0.0022$} & \multicolumn{1}{c|}{$0.132 \pm 0.0015$} \\ \hline
\multicolumn{1}{|c|}{BGOLEM}                              & \multicolumn{1}{c|}{$0.024 \pm 0.0002$} & \multicolumn{1}{c|}{$0.031 \pm 0.0003$}  \\ \hline
\end{tabular}
\end{table}
The PFDR metric compares the accuracy of each method in parts of the discovered ECs that are zero in the $\textrm{SC}_\textrm{B}$ and there is still no ground truth about the parts of ECs that are one in the $\textrm{SC}_\textrm{B}$., i.e., the PFDR metric is able to identify the part of the false positive in the ECs. We used the proportion test to compare the percentage of significantly different edges in ECs. In doing so, we searched for edges that have significantly different results by using different methods, namely FGES and GOLEM, in the population of $50$ subjects. Our results show that $40.3\%$ of EC edges that are zero in $\textrm{SC}_\textrm{B}$ are significantly different between these two methods, whereas this number is reduced to $11.2\%$ for the rest of the edges, suggesting that edges that are zero in $\textrm{SC}_\textrm{B}$ are more sensitive to the used method.

\begin{figure}[!htb]
\centering
\includegraphics[width=\textwidth]{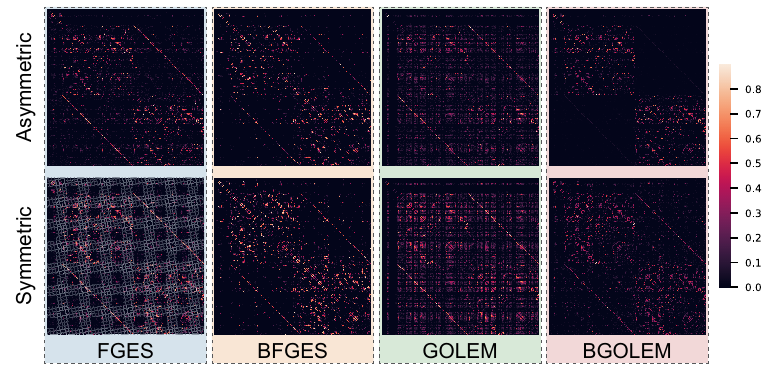}
\caption{Comparing the effect of employing Bayesian and non-Bayesian methods in the reproducibility of discovered asymmetric and symmetric ECs.
}
\label{fig:mesh7_001}
\end{figure}

To further investigate the effectiveness of the Bayesian causal frameworks on the \textbf{reliability} of discovered directed (asymmetric)  and undirected (symmetric) ECs, we use the Rogers-Tanimoto index \cite{rogers1960computer} that evaluates the reproducibility of each edge within the derived ECs. Note that, in undirected ECs, the connectome is primarily determined by the presence or absence of an edge, leading to a symmetric connectome. However, in directed ECs, both the existence or absence of edges and the direction of edges are discovered, leading to an asymmetric connectome. This index calculates the dissimilarity between the edges of the ECs obtained from the test and retest data of 50 subjects. By calculating this index for all the edges, we gain insights into the degree of alignment between the extracted ECs in the test and retest data. The resulting index value for each edge ranges from 0 to 1, where 0 indicates a perfect agreement and no dissimilarity of an edge in two sets of derived ECs, while 1 denotes no agreement and complete dissimilarity between an edge.
In Fig \ref{fig:mesh7_001}, we demonstrate the Rogers-Tanimoto values for all the edges of the ECs with 164 regions extracted using Bayesian and non-Bayesian methods for the test and retest data. The index is computed for both undirected and directed edges in this figure. 

\begin{figure}[!htb]
\centering
\includegraphics[width=\textwidth]{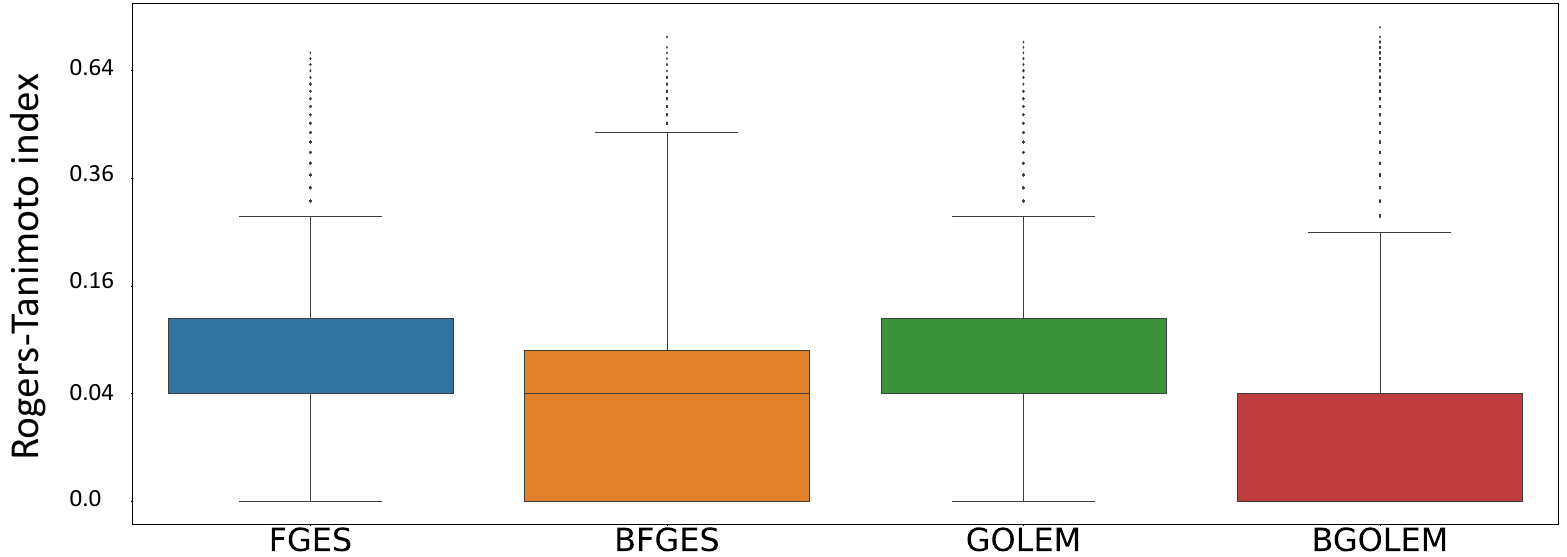}
\caption{The box plots of the Rogers-Tanimoto values of the FGES, BFGES, GOLEM, BGOLEM.
The median and interquartile range of the Rogers-Tanimoto index for directed ECs of the FGES method are $4\%$ and $[4, 11.5]\%$. These values for the directed ECs of the GOLEM methods are $4\%$ and $[4, 11.4]\%$. The median and interquartile range for the BFGES method are $4\%$ and $[0, 7.8]\%$, and for the BGOLEM method, these values are $0\%$ and $[0, 4]\%$.
}
\label{fig:mesh7}
\end{figure}

Fig \ref{fig:mesh7} shows the numerical values of the Rogers-Tanimoto index for all the directed edges discovered with the GOLEM, BGOLEM, FGES, and BFGES for test and retest data. 
The median and interquartile range of the Rogers-Tanimoto index for directed ECs of the FGES method are $4\%$ and $[4, 11.5]\%$. These values for the directed ECs of the GOLEM methods are $4\%$ and $[4, 11.4]\%$. The median and interquartile range for the BFGES method are $4\%$ and $[0, 7.8]\%$, and for the BGOLEM method, these values are $0\%$ and $[0, 4]\%$. According to Fig \ref{fig:mesh7}, the computed values of this index for the ECs of the Bayesian and non-Bayesian methods are highly right-skewed. To assess the significance of employing our Bayesian methods compared to the non-Bayesian method in terms of \textbf{reliability}, we conducted the Wilcoxon signed-rank test on the computed Roger-Tanimoto values. This test is performed due to its appropriation for the comparison of two sets of data measured on an ordinal scale that are highly skewed.
The p-value of this test obtained for comparing the Roger-Tanimoto values of BFGES and FGES is extremely small, with a value lower than 1e{-3}. Similarly, the p-value for comparing the Roger-Tanimoto values of BGOLEM and GOLEM is less than 1e{-3}. These results indicate that the observed differences in the Roger-Tanimoto values between BFGES and FGES, as well as between BGOLEM and GOLEM, are highly significant. The Rogers-Tanimoto index for undirected ECs is illustrated and discussed in Appendix 1.

\section*{Discussion}\label{4}
In this paper, first, the Bayesian causal frameworks are introduced in the ~\nameref{211} and ~\nameref{212} subsections. Then, the PFDR metric is presented in ~\nameref{22} subsection as a computable alternative for the FDR metric to assess the accuracy of causal discovery methods. 

Fig \ref{fig:mesh3_0} and \ref{fig:mesh3_001} show that the mean FDR and the mean Percent of Total Error are improved with the employment of Bayesian causal frameworks for SF7 data with different number of nodes. The results of applying the FGES and GOLEM methods to this data are consistent with the results in \cite{ng2020role}.
Fig \ref{fig:mesh3_1}, shows the relationship between FDR and PFDR using hybrid fMRI data with 164 nodes. The correlation coefficient between the PFDR and FDR of ECs obtained using the BGOLEM, GOLEM, BFGES, and FGES methods are close to one, i.e., the PFDR and FDR values are highly correlated in all of the methods. These results depict that the PFDR metric can be used as an accuracy metric instead of the FDR metric for discovering ECs where ground truth data is not available. It is worth noting that in the traditional method for evaluating the accuracy of causal discovery methods in extracting ECs, the first step involves collecting the differences in ECs discovered by applying a specific causal discovery method to both patient and normal subjects' data. Then, the aim is to assess the consistency between the discovered differences and the findings of neuroscientific studies. This type of evaluation is used in studies such as \cite{zhang2022detecting,rolls2018effective,rolls2020effective}. While this approach can be informative about some causal interactions in brain networks, the agreement/disagreement of small parts of ECs of patients and normal subjects does not necessarily imply the accuracy/inaccuracy of a causal method. For instance, an accurate functional connectome discovery method can identify a difference between two groups of data (patient and normal), however, identifying this difference does not imply that the derived connectomes are presenting the causal network of the brain.

Fig \ref{fig:mesh7_0} illustrates the structural and functional connectomes along with ECs discovered with Bayesian and non-Bayesian methods in various aspects. This figure investigates the visual symmetry of connections in the two hemispheres and the edges between them.
Comparison between the functional connectome derived from correlation values and the structural connectomes reveals that several regions are functionally highly correlated despite lacking structural connections. This functional connectome resembles those in \cite{sundaram2019individual}. Moreover, studies, such as \cite{Hellige2001Hemispheric,Toga2003Mapping,Corballis2014Left}, have extensively investigated brain asymmetry and provided insights into the functional differences between the two hemispheres. However, the functional connectome depicted in Fig \ref{fig:mesh7_0} exhibits high symmetry between the right and left hemispheres, which emphasizes of shortcomings of this method in understanding brain mechanisms. As a result, this contrast between the functional and structural connectomes underlines the opposition between the two, suggesting that the structural connectome can not serve as reliable prior knowledge when exploring functional connectivity. This is expected due to the existence of confounding nodes and the fact that functional connectivity refers to the statistical dependencies between different brain regions. These findings highlight the results presented in \cite{zalesky2012use} and \cite{goni2014resting}.

Fig \ref{fig:mesh7_0} helps to visually evaluate the potential of our results. From a structural perspective, it is well established that the two hemispheres of the brain are connected through commissural nerve tracts such as the Corpus callosum, which bridges the left and right hemispheres to share information.
According to the structural connectome presented in Fig \ref{fig:mesh7_0}, with the exclusion of subcortical regions of the brain, there is no structural direct edge between the two hemispheres. As a consequence, although we observe a high correlation between regions of two brain hemispheres, due to the satisfaction of the causal sufficiency assumption, it is not possible to establish a direct causal relationship between hemispheres.
In this figure, the ECs discovered through the FGES and GOLEM methods show connected nodes between the two hemispheres, with many edges having a low probability of existence in each hemisphere. Moreover, while the ECs discovered through effective methods can not be symmetrical between the two hemispheres, the ECs discovered with traditional causal discovery methods are highly symmetric, as it is shown in Fig \ref{fig:mesh7_0}. 
This emphasizes the limitations of discovering EC using state-of-the-art causal discovery methods like FGES and GOLEM, particularly when relying solely on fMRI data. These challenges can arise due to computational limitations (the weak performance of state-of-the-art methods in dealing with high-dimensional networks with short sample sizes) as well as practical constraints ( poor temporal resolution of fMRI data).
In contrast, the ECs discovered with Bayesian frameworks have a low number of edges between the two hemispheres, with a greater proportion of edges having a higher probability of existence in each hemisphere. These visual observations, along with the requirement that EC edges should be a subset of the edges in the SC, highlight the significance of employing Bayesian approaches in the investigation of EC.

To numerically evaluate the effectiveness of our method in the \textbf{accuracy} of derived ECs, we employed the PFDR metric and the results are presented in Fig \ref{fig:mesh6}.
Utilizing the GOLEM and FGES methods with empirical data shows that the GOLEM method has a lower mean PFDR value than that of the FGES method, which implies that the GOLEM method is more accurate in discovering EC (Fig \ref{fig:mesh6}.a).
Fig \ref{fig:mesh6}.b illustrates that both Bayesian methods decrease the PFDR values in EC discovery compared to the non-Bayesian methods. Considering the p-values of the Paired T-test, our numerical results imply that the proposed methods significantly decreased the errors in discovering accurate EC and enhanced the \textbf{accuracy} of discovering ECs in \cite{dubois2020causal} and \cite{zhang2022detecting}. Nevertheless, there is still no measure to assess the accuracy of causal discovery methods in determining the presence or absence of an edge in the EC, that is, one in the $\textrm{SC}_\textrm{B}$. The extracted ECs with different methods vary across subjects; however, this variance is more pronounced in areas of the EC whose corresponding edges are zero in $\textrm{SC}_\textrm{B}$. To support this statement, we used the proportion test to compare the percentage of significantly different edges in ECs when 200 edges are selected. In doing so, we searched for edges that had significantly different results by using different methods, namely FGES and GOLEM, in the population of $50$ subjects. Our results show that $40.3\%$ of edges that are not connected in $\textrm{SC}_\textrm{B}$ are significantly different between these two methods, whereas this number is reduced to $11.2\%$ for the rest of the edges, suggesting that edges that are not connected in $\textrm{SC}_\textrm{B}$ are more sensitive to the used method. Although we did this analysis for 200 edges in the ECs, similar results were obtained for other numbers of edges.

The \textbf{reliability} of EC discovery methods is numerically assessed using the Roger-Tanimoto index as a measure of reproducibility. Fig \ref{fig:mesh7_001} demonstrates that both directed and undirected edges between the two hemispheres exhibit greater similarity in the Bayesian methods, as indicated by their Roger-Tanimoto values. Furthermore, Fig \ref{fig:mesh7} illustrates a decrease in the Roger-Tanimoto index for the ECs when employing the Bayesian versions of the methods. The Bayesian causal frameworks exhibit significantly higher reproducibility (lower dissimilarities) in discovering directed ECs compared to the non-Bayesian FGES and GOLEM methods. 
Moreover, the BGOLEM method exhibits higher reproducibility compared to the BFGES method. The Roger-Tanimoto index values for the undirected ECs decrease when the Bayesian causal frameworks are employed, as shown in the appendix. According to the p-values of the Wilcoxon signed-rank tests, utilization of our Bayesian methods significantly improves the \textbf{reliability} of the discovered ECs compared to those derived in \cite{dubois2020causal} and \cite{zhang2022detecting} with the most state of the art methods, the FGES and GOLEM. It is important to note that the observed differences in the derived ECs using our Bayesian frameworks could be attributed to the intrinsic variations in brain connectivities among different subjects and not necessarily the weakness of causal discovery methods, as demonstrated in \cite{gong2011brain}.

\subsection*{Limitations and promising aspects of future research}
Future research needs to delve more deeply into four key limiting and challenging aspects. 
First, according to the findings of this paper, some errors that have already been demonstrated in the presented methods are caused by the limitations of  fMRI in presenting the brain activities and causal interactions in the brain \cite{logothetis2008we,plis2011effective}. 
As a result, in addition to DTI data, combining fMRI data with other neuroimaging techniques with a higher temporal resolution (such as EEG or MEG) can be beneficial in enhancing the accuracy of brain networks, as it is shown in \cite{jamadar2021metabolic}.
In this paper, we have shown the improvement of the discovered ECs by employing the Bayesian causal frameworks on both hybrid and real-world data. The degree of incorporation of the structural data in each method is optional, and various functions can be used to mask the prior information in each technique to discover a more accurate EC. However, there is no exact procedure for finding the optimal value that balances the impact of DTI and fMRI data on discovering ECs. This balancing value for fMRI and DTI fusion plays an essential role in the discovered ECs, and finding this value can be a second aspect of future research.

Moreover, considering multivariate Gaussian density distribution to estimate fMRI signals or assuming linear causal interaction between brain regions are fairly restrictive assumptions, as assumed in the FGES and GOLEM methods, respectively. 
As the third aspect of future works, since the activity in one neuronal population’s gates connection strengths among others, and this procedure causes highly nonlinear information transformation between brain regions, model-free methods need to be developed to measure these interactions and discover the EC. 

Most recent research investigates functional connectivity while considering time as one of the important parameters of their research, such as \cite{kim2021learning,shappell2019improved}, which can provide information about a possibly low-dimensional and intrinsic manifold of brain data, as it is mentioned in \cite{varley2022network}.
It is presented in \cite{vidaurre2017brain} that discovering dynamic brain networks can lead to a revolution in our understanding of brain functionality and the future promises of this research path are demonstrated in \cite{lurie2020questions}.
However, the existing research that is concerned with extracting effective connectivity from observational and interventional data is concerned with deriving static networks for brain EC and trying to interpret causal relations in the brain with them, and few research such as \cite{biswas2022statistical} have considered time as one the main component of EC discovery.
As the fourth and most important aspect of future studies, the next generation of EC discovery from observational and interventional data must derive dynamic effective networks (dynamic effective connectomes) to exploit temporal and spatial information. We believe this research path can lead to an even more profound understanding of brain mechanisms than dynamic brain networks.

\section*{Conclusion}
The ultimate goal of this paper is to develop more accurate and reliable methods for causal discovery in brain connectomes and to demonstrate their effectiveness both numerically and visually.
We introduce two Bayesian causal discovery methods that leverage DTI as prior knowledge. We then demonstrate the effectiveness of our methods through a series of simulations on synthetic and hybrid data. To assess the accuracy of the derived effective connectomes with our methods, we introduce the PFDR metric as a computational accuracy assessment. We show that our methods produce more accurate effective connectomes. Furthermore, we use the Roger Tanimoto index as a reproducibility metric to demonstrate that the effective connectomes of test-retest data derived from our methods are more reliable than those derived from traditional methods. Our study emphasizes the potential of these frameworks to significantly advance our understanding of brain function and organization.


%
%
%
\section*{Supporting information}


\paragraph*{S1 Appendix.}
\label{S2_Appendix}
{More details and discussions on our methods and results.}

\paragraph*{S2 Appendix.}
\label{S2_Appendix }
{The region names are provided in this Appendix.}

\end{document}